\documentstyle[lingmacros,twocolumn]{article}

\title{
Redundancy in Collaborative Dialogue \\
\begin{small} Appeared in Coling92\end{small}}
\author{
\begin{tabular}{cc}
Marilyn A. Walker\\
University of Pennsylvania, Computer Science Dept.\thanks{This research
was partially funded by ARO grant DAAL03-89-C0031PRI and DARPA grant
N00014-90-J-1863 at the University of Pennsylvania, by Hewlett
Packard, U.K., and by an NSF award for 1991 Summer Institute in Japan
}   \\
 Philadelphia, PA 19104  \\
lyn@linc.cis.upenn.edu
\end{tabular}
}

\date{}

\begin{document}           
\bibliographystyle{plain}  
\maketitle
\thispagestyle{empty}

\abstract
In dialogues in which both agents are autonomous, each agent
deliberates whether to accept or reject the contributions of the
current speaker. A speaker cannot simply assume that a proposal or an
assertion will be accepted. However, an examination of a corpus of
naturally-occurring problem-solving dialogues shows that agents often
do not explicitly indicate acceptance or rejection.  Rather the
speaker must infer whether the hearer understands and accepts the
current contribution based on indirect evidence provided by the
hearer's next dialogue contribution. In this paper, I propose a model
of the role of informationally redundant utterances in providing
evidence to support inferences about mutual understanding and
acceptance.  The model (1) requires a theory of mutual belief that
supports mutual beliefs of various strengths; (2) explains the
function of a class of informationally redundant utterances that
cannot be explained by other accounts; and (3) contributes to a theory
of dialogue by showing how mutual beliefs can be inferred in the
absence of the master-slave assumption.


\section{Introduction}
\label{intro-sec}

It seems a perfectly valid rule of conversation not to tell
people what they already know. Indeed, Grice's {\sc
quantity} maxim has often been interpreted this way: {\it
Do not make your contribution more informative than is
required}\cite{Grice67}.  Stalnaker, as well, suggests that
{\it to assert something that is already presupposed is to
attempt to do something that is already
done}\cite{Stalnaker78}.  Thus, the notion of what is {\it
informative} is judged against a background of what is {\it
presupposed}, i.e. propositions that all conversants assume
are mutually known or believed.  These propositions are
known as the {\sc common ground}\cite{Lewis69,Grice67}.

The various formulations of this `no redundancy' rule
permeate many computational analyses of natural language
and notions of cooperativity. However consider the
following excerpt from the middle of an advisory dialogue
between Harry (h), a talk show host, and Ray (r) his
caller\footnote{These examples come from the talk show for
financial advice, {\it Speaking of Your Money}, on WCAU in
Philadelphia. This corpus was collected and transcribed by
Martha Pollack and Julia Hirschberg\cite{PHW82}.}.

\begin{verbatim}
Example 1:
( 6) r. uh 2 tax questions.
     one: since April 81 we have had an
     85 year old mother living with us.
     her only income has been social security
     plus approximately $3000 from a
     certificate of deposit and i wonder
     what's the situation as far as
     claiming her as a dependent or does
     that income from the certificate of
     deposit rule her out as a dependent?
( 7) h. yes it does.
( 8) r. IT DOES.
( 9) h. YUP THAT KNOCKS HER OUT.
     .........
\end{verbatim}

In standard information theoretic terms, both (8) and (9)
are {\sc redundant}.  Harry's assertion in (9) simply
paraphrases what was said in (7) and (8) and so it cannot
be adding beliefs to the common ground\footnote{(8) is
not realized with a rising question intonation. This will
be discussed in section \ref{ind-ques-sec}.}. Furthermore, the
truth of (9) cannot be in question, for instead of (9),
Harry could not say {\it Yup, but that doesn't knock her
out}.  So why does Ray (r) in (8) {\sc repeat} Harry's (h)
assertion of {\it it does}, and why does Harry {\sc
paraphrase} himself and Ray in (9)?

My claim is that informationally redundant utterances
(IRU's) have two main discourse functions: (1) to provide
{\sc evidence} to support the assumptions underlying the
inference of mutual beliefs, (2) to {\sc center} a
proposition, ie. make or keep a proposition
salient\cite{GJW86}. This paper will focus on (1)
leaving (2) for future work.

First consider the notion of evidence.  One reason why
agents need {\sc evidence} for beliefs is that they only
have partial information about: (1) the state of world; (2)
the effects of actions; (3) other agent's beliefs,
preferences and goals. This is especially true when it
comes to modelling the effects of linguistic actions.
Linguistic actions are different than physical actions.  An
agent's prior beliefs, preferences and goals cannot be
ascertained by direct inspection. This means that it is
difficult for the speaker to verify when an action has
achieved its expected result, and so giving and receiving
evidence is critical and the process of establishing mutual
beliefs is carefully monitored by the conversants.

The characterization of IRU's as informationally redundant
follows from an axiomatization of action in dialogue that I
will call the {\sc deterministic model}.  This model
consists of a number of simplifying assumptions such as:
(1) Propositions are are either believed or not believed,
(2) Propositions representing beliefs and intentions get
added to the context by the unilateral action of one
conversant, (3) Agents are logically omniscient.  (4) The
context of a discourse is an undifferentiated set of
propositions with no specific relations between them. I
claim that these assumptions must be dropped in order to
explain the function of IRU's in dialogue.

Section \ref{bel-sec} discusses assumption (1); section
\ref{und-sec} shows how assumption (2) can be dropped;
section \ref{inf-sec} discusses (3); section
\ref{bel-rel-sec} shows that some IRU's facilitate the
inference of relations between adjacent propositions.

\section{Mutual Beliefs in a Shared Environment}
\label{bel-sec}

The account proposed here of how the {\sc common ground} is
augmented, is based is Lewis's {\sc shared environment}
model for common knowledge\cite{Lewis69,CM81}.  In this model,
mutual beliefs depend on evidence, openly available to the
conversants, plus a number of underlying
assumptions.

\begin{quote}
{\bf Shared Environment Mutual Belief Induction Schema} \\
It is mutually believed in a population
P that $\Psi$
if and only if some situation $\cal S$ holds such that:
\begin{enumerate}
\item Everyone in P has reason to believe that $\cal S$ holds.  
\item$\cal S$ indicates to everyone in P that everyone in P has reason to
       believe that $\cal S$  holds.  
\item$\cal S$ indicates to everyone in P that $\Psi$.
\end{enumerate}
\end{quote}

The situation $\cal S$, used above in the mutual belief
induction schema, is the context of what has been said.
This schema supports a weak model of mutual beliefs, that
is more akin to mutual assumptions or mutual
suppositions\cite{Prince78b}.  Mutual beliefs can be
inferred based on some evidence, but these beliefs may
depend on underlying assumptions that are easily
defeasible.  This model can be implemented using Gallier's
theory of autonomous belief revision and the corresponding
system\cite{Galliers91a}.

A key part of this model is that some types of evidence
provide better support for beliefs than other types.  The
types of evidence considered are categorized and ordered
based on the source of the evidence: {\tt hypothesis <
default < inference < linguistic < physical}(See
\cite{CM81,Galliers91a}).  This ordering reflects the {\bf
relative} defeasibility of different assumptions.
Augmenting the strength of an assumption thus decreases its
relative defeasibility.

A claim of this paper is that one role of IRU's is to
ensure that these assumptions are supported by evidence,
thus decreasing the defeasibility of the mutual beliefs
that depend on them\cite{Galliers91a}.

Thus mutual beliefs depend on a defeasible inference
process. All inferences depend on the evidence to support
them, and stronger evidence can defeat weaker evidence.  So
a mutual belief supported as an {\tt inference} can get
defeated by {\tt linguistic} information.  In addition, I
adopt an an assumption that a chain of reasoning is only as
strong as its weakest link:

\begin{quote}
{\bf Weakest Link Assumption}: The strength of a belief P
depending on a set of underlying assumptions $a_i,...a_n$
is MIN(Strength ($a_i,...a_n$))
\end{quote}

This seems intuitively plausible and means that the
strength of belief depends on the strength of underlying
assumptions, and that for all inference rules that depend
on multiple premises, the strength of an inferred belief is
the weakest of the supporting beliefs.


This representation of mutual belief differs from the
common representation in terms of an iterated
conjunction\cite{LA90} in that: (1) it relocates
information from mental states to the environment in which
utterances occur; (2) it allows one to represent the
different kinds of evidence for mutual belief; (3) it
controls reasoning when discrepancies in mutual beliefs are
discovered since evidence and assumptions can be inspected;
(4) it does not consist of an infinite list of statements.

\section{Inference of Understanding}
\label{und-sec}

This section examines the assumption from the {\sc
deterministic model} that: (2) Propositions representing
beliefs and intentions get added to the context by the
unilateral action of one conversant\footnote{This is an
utterance action version of the STRIPS assumption.}. This
assumption will also be examined in section
\ref{agree-sec}.

The key claim of this section is that agents monitor the
effects of their utterance actions and that the next action
by the addressee is taken as evidence of the effect of the
speaker's utterance\footnote{Except for circumstances where
it is clear that the flow of the conversation has been
interrupted.}. That the utterance will have the intended
effect is only a {\tt hypothesis} at the point where the
utterance has just been made, irrespective of the
intentions of the speaker. This distinguishes this account
from others that assume either that utterance actions
always succeed or that they succeed unless the addressee
previously believed otherwise\cite{LA90,GS90}.

I adopt the assumption that the participants in a dialogue
are trying to achieve some purpose\cite{GS86}.  Some
aspects of the structure of dialogue arises from the
structure of these purposes and their relation to one
another.  The minimal purpose of any dialogue is that an
utterance be understood, and this goal is a prerequisite to
achieving other goals in dialogue, such as commitment to
future action. Thus achieving mutual belief of
understanding is an instance of the type of activity that
agents must perform as they collaborate to achieve the
purposes of the dialogue. I claim that a model of the
achievement of mutual belief of understanding can be
extended to the achievement of other goals in dialogue.

\begin{figure*}[h,t]
\begin{center}
\begin{tabular}{|l|c|c|}
\hline & & \\
Next &  Assumption & Evidence  \\
Utterance & addressed & Type \\
\hline \hline & &\\
PROMPT  & attention & linguistic \\
\hline & & \\
REPEAT & hearing, attention & linguistic \\
\hline & & \\
PARAPHRASE  & realize, hearing, attention & linguistic \\
\hline & & \\
INFERENCE  & license, realize, hearing, attention & linguistic \\
\hline & & \\
IMPLICATURE  & license, realize, hearing, attention & linguistic \\
\hline \hline & & \\
ANY Next&    copresence  & linguistic \\
Utterance  &  license, realize, hearing, attention  & default \\
\hline
\end{tabular}
\caption{How the Addressee's Following utterance upgrades the evidence
underlying assumptions}
\label{ass-fig}
\end{center}
\end{figure*}

Achieving understanding is not unproblematic, it is a
process that must be managed, just as other goal achieving
processes are\cite{CS89}.  Inference of mutual
understanding relies upon some evidence, e.g. the utterance
that is made, and a number of underlying assumptions.  The
assumptions are given with the inference rule below.

\begin{verbatim}
say(A, B, u, p)  --A->
       understand(B, u, p) [evidence-type]

Assumptions =
{   copresent(A, B, u)     [evidence-type]
    attend(B, u)           [evidence-type]
    hear(B, u)             [evidence-type]
    bel(B, realize(u, p))  [evidence-type]
}
\end{verbatim}

This schema means that when A says u to B intending to
convey p, that this leads to the mutual belief that B
understands u as p under certain assumptions. The
assumptions are that A and B were copresent, that B was
attending to the utterance event, that B heard the
utterance, and that B believes that the utterance u
realizes the intended meaning p.

The {\tt [evidence-type]} annotation indicates the strength
of evidence supporting the assumption.  All of the
assumptions start out supported by no evidence; their
evidence type is therefore {\tt hypothesis}. It isn't until
{\bf after} the addressee's next action that an assumption
can have its strength modified.

The claim here is that one class of IRU's addresses these
assumptions underlying the inference of mutual
understanding. Each type of IRU, the assumption addressed
and the evidence type provided is given in Figure
\ref{ass-fig}. Examples are provided in sections \ref{rep-sec}
and \ref{para-sec}.

It is also possible that A intends that BY saying u, which
realizes p, B should make a certain inference q.  Then B's
understanding of u should include B making this inference.
This adds an additional assumption:

\begin{verbatim}
bel(B, license(p,q))  [evidence-type]
\end{verbatim}

Thus assuming that q was inferred relies on the
assumption that B believes that p licenses q in the
context.

Figure \ref{ass-fig} says that prompts, repetitions,
paraphrases and making inferences explicit all provide
linguistic evidence of attention.  All that prompts such as
{\it uh huh} do is provide evidence of attention.  However
repetitions, paraphrases and making inferences explicit
also demonstrate complete hearing.  In addition, a
paraphrase and making an inference explicit provides
linguistic evidence of what proposition the paraphraser
believes the previous utterance realizes.  Explicit
inferences additionally provide evidence of what inferences
the inferrer believes the realized proposition licenses in
this context.

In each case, the IRU addresses one or more assumptions
that have to be made in order to infer that mutual
understanding has actually been achieved.  The assumption,
rather than being a {\tt hypothesis} or a {\tt default},
get upgraded to a support type of {\tt linguistic} as a
result of the IRU.  The fact that different IRU's address
different assumptions leads to the perception that some IRU's
are better evidence for understanding than others, e.g. a
{\sc paraphrase} is stronger evidence of understanding than
a {\sc repeat}\cite{CS89}.

In addition, {\bf any} next utterance by the addressee can
upgrade the strength of the underlying assumptions to {\tt
default} (See Figure \ref{ass-fig}).  Of course {\tt
default} evidence is weaker than {\tt linguistic} evidence.
The basis for these default inferences will be discussed in
section \ref{agree-sec}.

\subsection{Example of a Repetition}
\label{rep-sec}

Consider example 1 in section \ref{intro-sec}.  Ray, in
(8), repeats Harry's assertion from (7).  This upgrades the
evidence for the assumptions of hearing and attention
associated with utterance (7) from {\tt hypothesis} to {\tt
linguistic}.  The assumption about what proposition p7 is
realized by u7 remains a {\tt default}.  This instantiates
the inference rule for understanding as follows:

\begin{verbatim}
 say(harry, ray, u7, p7) --A->
         understand(Ray, u7, p7) [default]

Assumptions =
{   copresent(harry, ray, u7) [linguistic]
    attend(ray, u7)           [linguistic]
    hear(ray, u7)             [linguistic]
    bel(ray, realize(u7, p7)) [default]
}
\end{verbatim}

Because of the {\sc weakest link} assumption, the belief
about understanding is still a default.

\subsection{Example of a Paraphrase}
\label{para-sec}

Consider the following excerpt:

\begin{verbatim}
Example 2:
(18) h. i see.  are there any other children
        beside your wife?
(19) d. no
(20) h. YOUR WIFE IS AN ONLY CHILD
(21) d. right. and uh wants to give
        her some security ..........
\end{verbatim}

Harry's utterance of (20) is said with a falling
intonational contour and hence is unlikely to be a
question. This utterance results in an instantiation of the
inference rule as follows:

\begin{verbatim}
 say(harry, ray, u20, p20) --A->
    understand(Ray, u20, p20) [linguistic]

Assumptions =
{   copresent(harry, ray, u7) [linguistic]
    attend(ray, u7)           [linguistic]
    hear(ray, u7)             [linguistic]
    bel(ray, realize(u7, p7)) [linguistic]
}
\end{verbatim}

In this case, the belief about understanding is supported
by {\tt linguistic} evidence since all of the supporting
assumptions are supported by linguistic evidence. Thus a
paraphrase provides excellent evidence that an agent
actually understood what another agent meant.

In addition, these IRU's leave a proposition salient, where
otherwise the discourse might have moved on to other
topics. This is part of the {\sc centering} function of
IRU's and is left to future work.

\section{Making Inferences Explicit}
\label{inf-sec}

This section discusses assumption (3) of the determistic
model, namely that: Agents are logically omniscient.  This
assumption is challenged by a number of cases in naturally
occurring dialogues where inferences that follow from what
has been said are made explicit.  I restrict the inferences
that I discuss to those that are (a) based on information
explicitly provided in the dialogue or, (b) licensed by
applications of Gricean Maxims such as scalar implicature
inferences\cite{Hirschberg85}.

For example the logical omniscience assumption would mean
that if \ex{1}(a) and (b) below are in the context, then
(c) will be as well since it is entailed from (a) and (b).

\eenumsentence
{\item[a.] You can buy an I R A
           if and only if you do NOT have an existing pension plan.
 \item[b.] You have  an existing pension plan.
 \item[c.] You cannot buy an I R A.
}

The following excerpt demonstrates this structure.
Utterance (15) realizes \ex{0}a, utterance (16) realizes
\ex{0}b, and utterance (17) makes the inference explicit
that is given in \ex{0}c for the particular tax year of
1981.

\begin{verbatim}
Example 3:
(15) h. oh no.
        I R A's were available
        as long as you are not a participant
        in an existing pension
(16) j. oh i see.
        well i did work i do work for a
        company that has a pension
(17) h. ahh. THEN YOU'RE NOT ELIGIBLE
        FOR EIGHTY ONE
(18) j. i see, but i am for 82
\end{verbatim}

After (16), since the propositional content of (17) is
inferrable, the assumption that Harry has made this inference
is supported by the {\tt inference} evidence type:

\begin{verbatim}
bel(H, license(p16, p17))  [inference]
\end{verbatim}

According to the model of achieving mutual understanding
that was outlined in section \ref{und-sec}, utterance (17)
provides {\tt linguistic} evidence that Harry (h) believes
that the proposition realized by utterance (16) licenses
the inference of (17) in this context.

\begin{verbatim}
bel(H, license(p16, p17))  [linguistic]
\end{verbatim}

Furthermore, the context here consists of a discussion of
two tax years {\it 1981} and {\it 1982}. Utterance (17)
selects {\it eighty one}, with a narrow focus pitch accent.
This implicates that there is some other tax year for which
Joe is eligible, namely {\it 1982}\cite{Hirschberg85}.
Joe's next utterance, {\it but I am for 82}, reinforces the
implicature that Harry makes in (17), and upgrades the
evidence underlying the assumption that (17) licenses (18)
to {\tt linguistic}.

\subsection{Supporting Inferences}
\label{bel-rel-sec}

A subcase of ensuring that certain inferences get made
involves the juxtaposition of two propositions.  These
cases challenge the assumption that: (4) The context of a
discourse is an undifferentiated set of propositions with
no specific relations between them. While this assumption
is certainly not made in most discourse models, it is often
made in semantic models of the context\cite{Stalnaker78}.
In the following segment, Jane (j) describes her financial
situation to Harry (h) and a choice between a settlement
and an annuity.

\begin{verbatim}
Example 4:
( 1) j. hello harry, my name is jane
( 2) h. welcome jane
( 3) j. i just retired december first,
     and in addition to my pension and
     social security, I have a
     supplemental annuity
( 4) h. yes
( 5) j. which i contributed to
        while i was employed
( 6) h. right
( 7) j. from the state of NJ mutual fund.
     and I'm entitled to a lump sum
     settlement which would be between
     16,800 and 17,800, or a lesser life
     annuity. and the choices of the annuity
     um would be $125.45 per month.
     That would be the maximum
     with no beneficiaries
( 8) h. You can stop right there:
        take your money.
( 9) j. take the money.
(10) h. absolutely.
        YOU'RE ONLY GETTING 1500 A YEAR.
        at 17,000, no trouble at all to
        get 10 percent on 17,000 bucks.
\end{verbatim}

Harry interrupts her at (8) since he believes he has enough
information to suggest a course of action, and tells her
{\it take your money}.  To provide {\sc support} for this
course of action he produces an inference that follows from
what she has told him in (7), namely {\it You're only
getting 1500 (dollars) a year}. {\sc support} is a general
relation that holds between beliefs and intentions in this
model.

Presumably Jane would have no trouble calculating that
\$125.45 a month for 12 months amounts to a little over
\$1500 a year, and thus can easily accept this statement
that is intended to provide the necessary {\sc support}
relation, ie. the juxtaposition of this fact against the
advice to {\it take the money} conveys that the fact that
she is only getting 1500 dollars a year is a reason for her
to adopt the goal of taking the money, although this is not
explicitly stated.


\section{Evidence of Acceptance}
\label{agree-sec}

In section \ref{und-sec}, I examine the assumption that: (2)
Propositions representing beliefs and intentions get added
to the context by the unilateral action of one conversant.
I suggested that this assumption can be replaced by
adopting a model in which agents' behavior provides
evidence for whether or not mutual understanding has been
achieved.  I also discussed some of the effects of resource
bounds, ie. cases of ensuring that or providing evidence
that certain inferences dependent on what is said are made.

Achieving understanding and compensating for resource
bounds are issues for a model of dialogue whether or not
agents are autonomous. But agents' autonomy means there are
a number of other reasons why A's utterance to B conveying
a proposition p might not achieve its intended effect: (1)
p may not cohere with B's beliefs, (2) B may not think that
p is relevant, (3) B may believe that p does not contribute
to the common goal, (4) B may prefer doing or believing
some q where p is mutually exclusive with q, (5) If p is
about an action, B may want to partially modify p with
additional constraints about how, or when p.

Therefore it is important to distinguish an agent actually
{\sc accepting} the belief that p or intending to perform
an action described by p from merely understanding that p
was conveyed.  Other accounts legislate that helpful agents
should adopt other's beliefs and intentions or that
acceptance depends on whether or not the agent previously
believed $\neg$ p\cite{LA90,GS90}.  But agents can decide
whether as well as how to revise their
beliefs\cite{Galliers91a}.

Evidence of acceptance may be given explicitly, but
acceptance can be inferred in some dialogue situations via
the operation of a simple principle of cooperative
dialogue\footnote{This is a simplification of the {\sc
collaborative planning principles} described in
\cite{WW90}.}:

\begin{quote}
{\sc collaborative principle}: Conversants must provide
evidence of a detected discrepancy in belief as soon as
possible.
\end{quote}

This principle claims that evidence of conflict should be
made apparent in order to keep {\tt default} inferences
about acceptance or understanding from going through.
IRU's such as {\sc prompts}, {\sc repetitions}, {\sc
paraphrases}, and making an {\sc inference} explicit cannot
function as evidence for conflicts in beliefs or intentions
via their propositional content since they are
informationally redundant. If they are realized with
question intonation, the inference of acceptance is
blocked.

In the dialogue below between Harry (h) and Ruth (r), Ruth
in (39), first ensures that she understood Harry correctly,
and then provides explicit evidence of non-acceptance in
(41), based on her autonomous preferences about how her
money is invested. .

\begin{verbatim}
Example 5:
(38) h. and I'd like 15 thousand in a
        2 and a half year certificate
(39) r. the full 15 in a 2 and a half?
(40) h. that's correct
(41) r. GEE. NOT AT MY AGE
\end{verbatim}

In the
following example, Joe in (14) makes a statement that
provides propositional content that conflicts with Harry's
statement in (13) and thus provides evidence of
non-acceptance.

\begin{verbatim}
Example 6
(13) h. and  -- there's no reason why you
    shouldn't have an I R A  for last year
(14) j. WELL I THOUGHT THEY JUST STARTED
     THIS YEAR
\end{verbatim}

Joe's statement is based on his prior beliefs.  In both of
these cases this evidence for conflict is given
immediately.  However when there is no evidence to the
contrary\footnote{This displaying of evidence to the
contrary was called an interruption in \cite{WW90}.}, and
goals of the discourse require achievement of acceptance,
inferences about acceptance are licensed as {\tt default}.
They can be defeated later by stronger evidence.

Without this principle, a conversant might not bring up an
objection until much later in the conversation, at which
point the relevant belief and some inferences following
from that belief will have been added to the common ground
as {\tt defaults}. The result of this is that the
retraction of that belief results in many beliefs being
revised.  The operation of this principle helps conversants
avoid replanning resulting from inconsistency in beliefs,
and thus provides a way to manage the augmentation of the
common ground efficiently.

\section{Other hypotheses}

The first point to note is that the examples here are only
a subset of the types of IRU's that occur in dialogues.  I
use the term antecedent to refer to the most recent
utterance which should have added the proposition to the
context.  This paper has mainly focused on cases where the
IRU: (1) is adjacent to its antecedent, rather than remote;
(2) realizes a proposition whose antecedent was said by
another conversant, (3) has only one antecedent. It is with
respect to this subset of the data that the alternate
hypotheses are examined.

A distributional analysis of a subset of the corpus (171
IRU's from 24 dialogues consisting of 976 turns), on the
relation of an IRU to its antecedent and the context, shows
that 35\% of the tokens occur remotely from their
antecedents, that 32\% have more than one antecedent, that
48\% consist of the speaker repeating something that he
said before and 52\% consist of the speaker repeating
something that the other conversant said. So the data that
this paper focuses on accounts for about 30\% of the data.

\subsection{Indirect Question Hypothesis}
\label{ind-ques-sec}

In example (1) of section \ref{intro-sec}, an alternative
account of Ray's repetition in (8) is that it is a question
of some kind.  This raises a number of issues: (1) Why
doesn't it have the form of a question?, (2) What is it a
question about?, and (3) Why is it never denied?.

Of 171 IRU's, only 28 are realized with rising question
intonation. Of these 28, 6 are actually redundant questions
with question syntax, and 14 are followed by affirmations.

If these are generally questions, then one possible answer
to what the question is about is that Ray is questioning
whether he actually heard properly. But then why doesn't he
use an intonational contour that conveys this fact as Ruth
does in example 5? On an efficiency argument, it is hard to
imagine that it would have cost Ray any more effort to have
done so.

Finally, if it were a question it would seem that it should
have more than one answer. While 50 of these IRU's are
followed by an affirmation such as {\it that's correct,
right, yup}, none of them are ever followed by a denial of
their content. It seems an odd question that only has one
answer.

\subsection{Dead Air Hypothesis}

Another hypothesis is that IRU's result from the radio talk
show environment in which silence is not tolerated.  So
agents produce IRU's because they cannot think of anything
else to say but feel as though they must say something.

The first point to note is that IRU's actually occur in
dialogues that aren't on the radio\cite{Carletta92}.  The
second question is why an agent would produce an IRU,
rather than some other trivial statement such as {\it I
didn't know that}. Third, why don't these utterance
correlate with typical stalling behavior such as false
starts, pauses, and filled pauses such as {\it uhhh}.

The dead air hypothesis would seem to rely on an assumption
that at unpredictable intervals, agents just can't think
very well.  My claim is that IRU's are related to goals,
that they support inferencing and address assumptions
underlying mutual beliefs, ie. they are not random.  In
order to prove this it must be possible to test the
hypothesis that it is only {\bf important} propositions
that get repeated, paraphrased or made explicit. This can
be based on analyzing when the information that is repeated
has been specifically requested, such as in the caller's
opening question or by a request for information from
Harry. It should also be possible to test whether the IRU
realizes a proposition that plays a role in the final plan
that Harry and the caller negotiate.  However this type of
strong evidence against the dead air hypothesis is left to
future work.

\section{Discussion}

It should be apparent from the account that the types of
utterances examined here are not really redundant. The
reason that many models of belief transfer in dialogue
would characterize them as redundant follows from a
combination of facts: (1) The representation of belief in
these models has been binary; (2) The effects of utterance
actions are either assumed to always hold, or to hold as
defaults unless the listener already believed otherwise.
This means that these accounts cannot represent the fact
that a belief must be supported by some kind of evidence
and that the evidence may be stronger or weaker. It also
follows from (2) that these models assume that agents are
not autonomous, or at least do not have control over their
own mental states. But belief revision is surely an
autonomous process; agents can choose whether to accept a
new belief or revise old
beliefs\cite{Galliers91a,GS90}.

The occurrence of IRU's in dialogue has many ramifications
for a model of dialogue. Accounting for IRU's has two
direct effects on a dialogue model. First it requires a
model of mutual beliefs that specifies how mutual beliefs
are inferred and how some mutual beliefs can be as weak as
mutual suppositions.  One function of IRU's is to address
the assumptions on which mutual beliefs are based. Second
the assumption that propositions representing beliefs and
intentions get added to the context by the unilateral
action of one conversant must be dropped.  This account
replaces that assumption with a model in which the evidence
of the hearer must be considered to establish mutual
beliefs. The claim here is that both understanding and
acceptance are monitored.  The model outlined here can be
used for different types of dialogue, including dialogues
in which agents are constructing mutual beliefs to support
future action by them jointly or alone.

How and when agents decide to augment the strength of
evidence for a belief has not been addressed in this work
as yet. Future work includes analyzing the corpus with
respect to whether the IRU plays a role in the final plan
that is negotiated between the conversants.

\section{Acknowledgements}

Discussions with Aravind Joshi, Ellen Prince and Bonnie
Webber have been extremely helpful in the development of
these ideas.  In addition I would like to thank Herb Clark,
Sharon Cote, Julia Galliers, Ellen Germain, Beth Ann
Hockey, Megan Moser, Hideyuki Nakashima, Owen Rambow,
Craige Roberts, Phil Stenton, and Steve Whittaker for the
influence of their ideas and for useful discussions.


\end{document}